\documentclass[10pt,twocolumn,letterpaper]{article}

\usepackage[pagenumbers]{iccv} 

\definecolor{iccvblue}{rgb}{0.21,0.49,0.74}
\usepackage[pagebackref,breaklinks,colorlinks,allcolors=iccvblue]{hyperref}
\usepackage{amsmath,amssymb,amsfonts}
\usepackage{algorithmic}
\usepackage{graphicx}
\usepackage{textcomp}
\usepackage{xcolor}
\usepackage{upgreek}
\usepackage{multirow}
\usepackage{booktabs}
\usepackage{makecell}

\def\paperID{*****} 
\def\confName{ICCV}
\def\confYear{2025}

\title{GI-SLAM: Gaussian-Inertial SLAM}

\author{Xulang Liu\\
Sun Yat-sen University\\
Guangzhou, China\\
\and
Ning Tan\\
Sun Yat-sen University\\
Guangzhou, China\\
}

\begin{document}
\maketitle
\begin{abstract}
3D Gaussian Splatting (3DGS) has recently emerged as a powerful representation of geometry and appearance for dense Simultaneous Localization and Mapping (SLAM). Through rapid, differentiable rasterization of 3D Gaussians, many 3DGS SLAM methods achieve near real-time rendering and accelerated training. However, these methods largely overlook inertial data, witch is a critical piece of information collected from the inertial measurement unit (IMU). In this paper, we present GI-SLAM, a novel gaussian-inertial SLAM system which consists of an IMU-enhanced camera tracking module and a realistic 3D Gaussian-based scene representation for mapping. Our method introduces an IMU loss that seamlessly integrates into the deep learning framework underpinning 3D Gaussian Splatting SLAM, effectively enhancing the accuracy, robustness and efficiency of camera tracking. Moreover, our SLAM system supports a wide range of sensor configurations, including monocular, stereo, and RGBD cameras, both with and without IMU integration. Our method achieves competitive performance compared with existing state-of-the-art real-time methods on the EuRoC and TUM-RGBD datasets.
\end{abstract}
\section{Introduction}
\label{sec:intro}

Dense Visual Simultaneous Localization and Mapping (SLAM) is a fundamental problem in 3D computer vision with numerous applications in fields such as autonomous driving, robotic navigation and planning, as well as augmented reality (AR) and virtual reality (VR). The goal of dense SLAM is to construct a detailed map of an unknown environment while simultaneously tracking the camera pose, ensuring that embodied agents can perform downstream tasks in previously unseen 3D environments. Traditional SLAM methods utilize point clouds\cite{10.1145/2030112.2030123,7946260,9010395}, surfels\cite{8794101}, meshes\cite{9009776}, or voxel grids\cite{9197357} for scene representation to create dense maps, achieving significant progress in localization accuracy. However, these methods still face substantial limitations for applications requiring high-fidelity, fine-grained 3D dense mapping.

To enhance the visual fidelity of 3D maps, researchers have explored SLAM methods based on Neural Radiance Fields (NeRF)\cite{NeRF10.1145/3503250}. These approaches leverage differentiable volumetric rendering to train implicit neural networks, enabling photometrically accurate 3D representations of the environment. This novel map representation is compact, continuous, and efficient, and it can be optimized through differentiable rendering, making it particularly beneficial for applications such as navigation, planning, and reconstruction. Methods such as iMAP\cite{9710431} pioneered the use of MLP-based representations for real-time tracking and mapping, while NICE-SLAM\cite{NICE-SLAM9878912} introduced a hierarchical voxel-based strategy to enhance scalability and reconstruction quality. Further advancements like Vox-Fusion\cite{Vox-Fusion9995035} and ESLAM\cite{10205103} have improved memory efficiency and tracking robustness through dynamic voxel allocation and multi-scale feature planes. Co-SLAM\cite{10204198} explore InstantNGP\cite{10.1145/3528223.3530127} to further accelerate the mapping speed. Despite these advancements, NeRF-SLAM still faces challenges such as high computational cost, slow convergence, and difficulties in handling large-scale dynamic environments due to model capacity limitations and catastrophic forgetting.

More recently, 3D Gaussians\cite{3DGS10.1145/3592433} have emerged as an alternative representation of radiance fields, achieving equal or superior rendering quality compared to traditional NeRFs while being significantly faster in rendering and training. Building on this success, some works\cite{SplaTAM10656349,Photo-SLAM10657868,GSFusion10758260,GS-SLAM10657581,MonoGS10657715} have integrated 3D Gaussian Splatting (3DGS) with dense visual SLAM systems and achieved significant advancements in both mapping accuracy and computational efficiency. However, to the best of our knowledge, no existing work has effectively integrated inertial data—which is crucial for accurate camera pose estimation in visual SLAM—into a 3DGS SLAM framework.

IMUs are commonly found in many mobile devices and robots, and they are frequently integrated into traditional visual SLAM systems\cite{VINS-Mono8421746,ORBSLAM3-9440682,10160620,10342106,9981134} to enhance the accuracy and robustness of camera tracking. To this end, we propose GI-SLAM, a novel Gaussian-inertial SLAM system that consists of an IMU-enhanced camera tracking module and a photo-realistic 3D Gaussian scene representation for mapping. In our system, we introduce a novel IMU loss function, which, when combined with the photometric loss function, improves the accuracy and robustness of camera tracking. Additionally, we propose a keyframe selection strategy based on Gaussian visibility and motion data to prevent the selection of motion-blurred keyframes, thereby improving the quality of the reconstructed map.
In summary, the main contributions of this work include:
\begin{itemize}
\item We propose a novel simultaneous localization and photorealistic mapping system based on 3D Gaussian Splatting (3DGS), called GI-SLAM. The proposed framework supports monocular, stereo, and RGBD cameras, with or without IMU integration, and is designed to operate effectively in both indoor and outdoor environments.
\item We present a keyframe selection strategy based on Gaussian visibility and motion constraint to prevent the selection of motion-blurred keyframes.
\item We conducted extensive evaluations of our proposed system on the TUM and EuRoC datasets with IMU data, demonstrating competitive performance.
\end{itemize}

\begin{figure*}[ht!]
    \centerline{\includegraphics[width=1.0\textwidth]{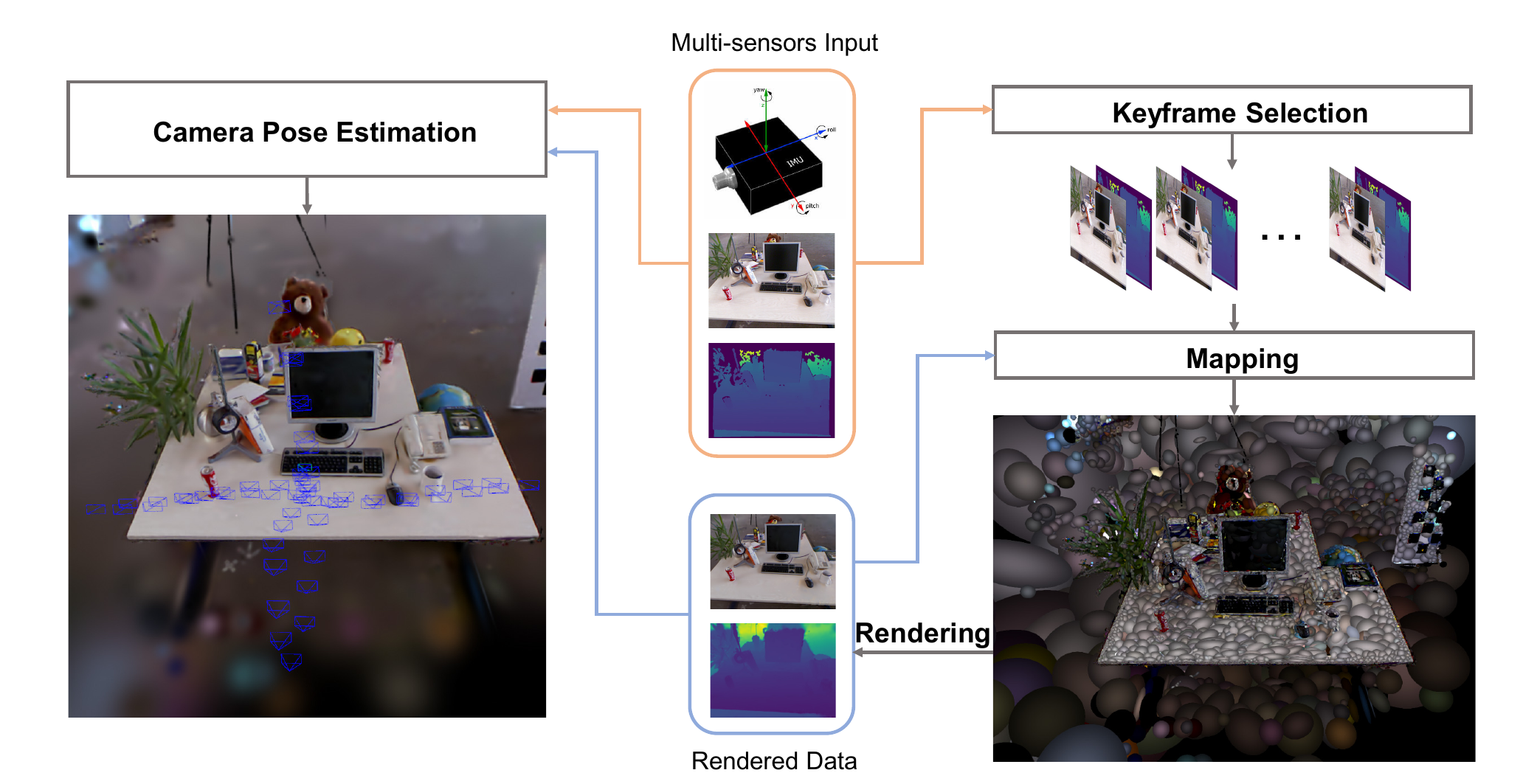}}
    \caption{SLAM system overview.}
    \label{fig1}
\end{figure*}

\section{Related Work}
\label{sec:related}

In this section, we provide a concise overview of various dense SLAM approaches, with a particular focus on recent methods that utilize implicit representations encoded in overfit neural networks for tracking and mapping. For a more comprehensive review of NeRF-based and 3DGS-based SLAM techniques, we refer interested readers to the excellent survey\cite{tosi2024nerfs}.

\paragraph{Classical dense SLAM}
There has been extensive work on 3D reconstruction over the past decades. Traditional dense SLAM methods have explored various map representations, including point clouds\cite{10.1145/2030112.2030123,6599048,6696650}, surfels\cite{Whelan-RSS-15,8794101,8954208}, and Gaussian mixture models\cite{10295571,10068771}. truncated signed-distance functions (TSDF)\cite{6162880,10.1145/2461912.2461940,10.1145/3072959.3054739,9662197}. Among these, Surfel-based approaches like ElasticFusion\cite{Whelan-RSS-15} represent the scene as a collection of circular surface elements, enabling real-time tracking and mapping. TSDF-based methods such as KinectFusion\cite{6162880} and its extensions\cite{10.1145/2508363.2508374,8202315} have demonstrated impressive real-time 3D reconstruction capabilities by leveraging efficient volumetric integration and voxel hashing techniques. While these classical methods provide geometrically accurate reconstructions, they often struggle with scalability and handling uncertainty in depth measurements. Recent advances have introduced learning-based methods such as RoutedFusion\cite{9156953} and DI-Fusion\cite{9577541}, which leverage deep networks to improve robustness against noisy depth inputs. These works primarily focus on the geometry reconstruction, while differently, our method takes both 3D reconstruction and photorealistic rendering into account simultaneously.

\paragraph{NeRF-based dense SLAM}
Several recent works have explored the use of neural radiance fields(NeRF)\cite{NeRF10.1145/3503250} for dense SLAM, leveraging its ability to efficiently represent scene geometry and appearance. For instance, iMAP\cite{9710431} pioneered the integration of neural implicit representations into SLAM, using a single MLP to dynamically build a scene-specific 3D model. This approach provides efficient geometry representation and automatic detail control but struggles with scalability and catastrophic forgetting in larger environments. NICE-SLAM\cite{NICE-SLAM9878912} improved upon iMAP by adopting a hierarchical feature grid representation with multi-level voxel encodings, enhancing reconstruction quality and mitigating catastrophic forgetting through localized updates. Vox-Fusion\cite{Vox-Fusion9995035} further introduced an octree-based voxel allocation strategy, allowing for dynamic scene encoding and efficient memory management. More recent advancements, such as ESLAM\cite{10205103}, utilize multi-scale feature planes to optimize memory efficiency and improve reconstruction speed by leveraging a TSDF-based representation. Co-SLAM\cite{10204198} combines smooth coordinate encodings with sparse hash grids, achieving robust tracking and high-fidelity map reconstruction with efficient hole-filling techniques. Additionally, GO-SLAM\cite{10378579} integrates global optimization techniques like loop closure and bundle adjustment to ensure long-term trajectory consistency. Alternative representations have also been explored. Point-SLAM\cite{Point-SLAM10377819} employs dynamic neural point clouds, adaptively adjusting density based on scene information for memory-efficient mapping. Other works such as Plenoxel-SLAM\cite{10484344} eschew neural networks entirely, leveraging voxel grids with trilinear interpolation for efficient real-time mapping and tracking. Despite these advances, NeRF-based SLAM systems face challenges related to computational complexity and memory consumption, which hinder real-time performance on large-scale scenes. Our 3DGS-based approach aims to address these limitations by introducing a more memory-efficient representation and faster rendering pipeline.

\paragraph{3DGS-based dense SLAM}
 Recently, with the great success of 3D Gaussian Splatting(3DGS)\cite{3DGS10.1145/3592433}, some works have integrated 3DGS with dense SLAM systems. MonoGS\cite{MonoGS10657715}  employs 3D Gaussians as the sole representation for online reconstruction, utilizing direct optimization for camera tracking and introducing Gaussian shape regularization to maintain geometric consistency. Photo-SLAM\cite{Photo-SLAM10657868} integrates explicit geometric features with implicit texture representations within a hyper primitives map, optimizing camera poses through multi-threaded factor graph solving. SplaTAM\cite{SplaTAM10656349} refines tracking and mapping by optimizing Gaussian parameters through re-rendering errors and incremental densification strategies, though it remains sensitive to motion blur and depth noise. GS-SLAM\cite{GS-SLAM10657581} introduces an adaptive expansion strategy to dynamically manage 3D Gaussians and a coarse-to-fine tracking approach to enhance pose estimation accuracy. Gaussian-SLAM\cite{yugay2023gaussianslam} organizes scenes into independently optimized sub-maps based on camera motion, improving scalability for larger environments. Compact-GSSLAM\cite{10655367} addresses memory constraints by reducing Gaussian ellipsoid parameters and employing a sliding window-based masking strategy for efficient resource allocation. Despite these advancements, challenges persist in achieving robust real-time performance, particularly in handling dynamic scenes and motion blur. Our approach introduces an IMU-based loss to incorporate inertial information into the 3DGS framework, significantly enhancing camera tracking accuracy and robustness. Additionally, motion constraints are utilized to filter frames susceptible to motion blur, thereby improving overall mapping quality. This integration of IMU data with 3DGS-based SLAM ensures more stable tracking and higher-fidelity reconstructions in challenging environments.

\section{Method}
\label{sec:method}

GI-SLAM contains three main components, including localization, mapping, and keyframing, shown in \mbox{\cref{fig1}}. Our proposed framework is an online SLAM system that simultaneously tracks camera poses and reconstructs dense scene geometry using RGB, depth, and IMU data. This is achieved through the following steps: First, the camera pose is initialized, and an initial 3D Gaussian scene representation is generated. Camera tracking is then performed by comparing the rendered RGB and depth images from the 3D Gaussian map with the input images, depth data, and IMU measurements. Each incoming frame is evaluated to determine whether it qualifies as a keyframe. If identified as a keyframe, the 3D Gaussian map is updated accordingly, which in turn enhances subsequent camera tracking. This process repeats iteratively until the SLAM system is terminated.

\begin{figure*}[h!]
    \centerline{\includegraphics[width=0.9\textwidth]{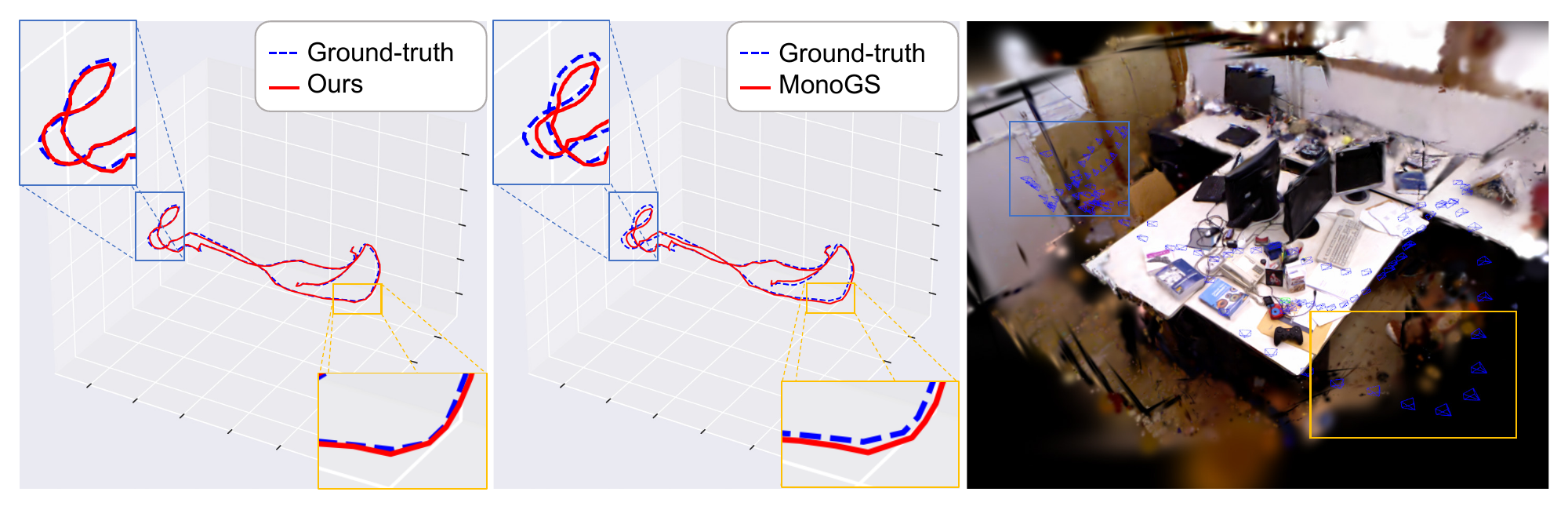}}
    \caption{Trajectory tracking result of the monocular camera setup on the TUM dataset.}
    \label{fig2}
\end{figure*}

\subsection{Localization}
We take advantage of the differentiable nature of the 3D Gaussian map rendering process by parameterizing the pose matrix used to represent the camera pose, enabling optimization via gradient descent. The camera pose is represented using an $SE(3)$ homogeneous transformation matrix:
\begin{equation}
    \mathbf{P}_t\in SE(3)=
        \begin{bmatrix}
            \mathbf{R}_t & \mathbf{t}_t \\
            \mathbf{0}^\top & 1
        \end{bmatrix} .
\end{equation}

We assume that the previous camera pose is accurate, and update the current camera pose based on it: 
\begin{equation}
    \mathbf{P}_t = \mathbf{P}_{t-1} \cdot \Delta \mathbf{P},
\end{equation}
where $\mathbf{P}_t$ represents the current camera pose, $\mathbf{P}_{t-1} $ denotes the previous pose, and $\Delta \mathbf{P}$ is the incremental transformation estimated through optimization. The refinement of $\Delta \mathbf{P}$ is achieved by minimizing the alignment error between the sensor data (RGB, depth, and IMU measurements) and the corresponding RGB and depth images rendered from the 3D Gaussian map. This optimization process involves minimizing a weighted combination of three loss functions: the RGB loss, which measures photometric consistency; the depth loss, which ensures geometric accuracy; and the IMU loss, which enforces inertial constraints.

\paragraph{RGB Loss}
In the monocular case, we minimise the following RGB loss:
\begin{equation}
    \mathcal{L}_{rgb}=\left\|I(\mathcal{G},\mathbf{P}_{t-1} \cdot \Delta \mathbf{P})-\bar{I}\right\|_1 .
\end{equation}
This is an L1 loss function, where $I(\mathcal{G},\mathbf{P}_{t-1} \cdot \Delta \mathbf{P})$
represents the process of rendering the 3D Gaussians $\mathcal{G}$ into a 2D image at pose $\mathbf{P}_{t-1} \cdot \Delta \mathbf{P}$, and $\bar{I}$ denotes the ground truth image.

\paragraph{Depth Loss}
When depth observations are available, we define the depth loss as:
\begin{equation}
    \mathcal{L}_{depth}=\left\|D(\mathcal{G},\mathbf{P}_{t-1} \cdot \Delta \mathbf{P})-\bar{D}\right\|_1,
\end{equation}
where $D(\mathcal{G},\mathbf{P}_{t-1} \cdot \Delta \mathbf{P})$
represents depth rasterisation, and $\bar{D}$ denotes the observed depth.

\paragraph{IMU Loss}
Our IMU loss function incorporates both translational and rotational constraints from 6-DOF IMU measurements. For translational constraints, we integrate the linear acceleration measurements $\mathbf{a}_t$ with the previous camera frame's linear velocity $\mathbf{v}_{t-1}$:

\begin{equation}
    \Delta\mathbf{p}_{imu} = \mathbf{v}_{t-1}\Delta t + \frac{1}{2}\mathbf{a}_t\Delta t^2.
\end{equation}
The translation loss $\mathcal{L}_{trans}$ is then computed as:  
\begin{equation}
    \mathcal{L}_{trans} = \|\Delta\mathbf{p}_{opt} - \Delta\mathbf{p}_{imu}\|^2_2,
\end{equation}
where $\Delta\mathbf{p}_{opt} \in \mathbb{R}^3$ denotes the optimized displacement between consecutive frames.

For rotational constraints, we derive the relative rotation from angular velocity measurements $\mathbf{\omega}_t$:
\begin{equation}
    \Delta\mathbf{\uptheta}_{imu} = \mathbf{\omega}_t \Delta t.
\end{equation}
The rotation loss $\mathcal{L}_{rot}$ is formulated as:  
\begin{equation}
    \mathcal{L}_{rot} = \|\Delta\mathbf{\uptheta}_{opt} - \Delta\mathbf{\uptheta}_{imu}\|^2_2,
\end{equation}
where $\Delta\mathbf{\uptheta}_{opt} \in \mathbb{R}^3$ represents the optimized relative rotation in axis-angle form.
The final IMU loss combines both components through weighted summation: 
\begin{equation}
    \mathcal{L}_{imu} = \lambda_t\mathcal{L}_{trans} + \lambda_r\mathcal{L}_{rot},
\end{equation}
with $\lambda_t$ and $\lambda_r$ being hyper-parameters balancing the two constraints. Both $\Delta\mathbf{p}_{opt}$ and $\Delta\mathbf{\uptheta}_{opt}$ are optimized variables in our SLAM framework.

\subsection{Mapping}
Our method employs 3D Gaussians as the sole 3D scene representation to model the environment, capable of rendering precise RGB and depth images through differentiable rendering. This facilitates the use of gradient-based optimization techniques to update the 3D environmental map.

\paragraph{3D Gaussian Scene Representation}
Similar to MonoGS\cite{MonoGS10657715}, our SLAM system models the scene as a collection of anisotropic 3D Gaussians $\mathcal{G} = \{G_i\}$, where each Gaussian $G_i$ is parameterized by its position $\boldsymbol{\mu}_W^i \in \mathbb{R}^3$, covariance $\boldsymbol{\Sigma}_W^i \in \mathbb{R}^{3\times3}$ (defining ellipsoidal geometry), color $c^i \in \mathbb{R}^3$, and opacity $\alpha^i \in \mathbb{R}$. To balance expressiveness and computational efficiency, we parameterize the covariance matrix as:  
\begin{equation}
    \boldsymbol{\Sigma}_W^i = \mathbf{R}_i \mathbf{S}_i \mathbf{S}_i^\top \mathbf{R}_i^\top,
\end{equation}
where $\mathbf{S}_i \in \mathbb{R}^3$ is a diagonal scaling matrix, and $\mathbf{R}_i \in \text{SO}(3)$ (represented as a quaternion) defines rotation. This formulation avoids explicit surface extraction and enables efficient scene reconstruction through volume rendering.  The continuous 3D representation allows adaptive refinement of Gaussians during mapping.

\paragraph{Differentiable Rendering}
To bridge 3D Gaussians with 2D observations, we project $G_i$ onto the image plane via a differentiable geometric transformation. Given camera pose $\boldsymbol{T}_{CW} \in \text{SE}(3)$, the 3D Gaussian $\mathcal{N}(\boldsymbol{\mu}_W, \boldsymbol{\Sigma}_W)$ maps to a 2D Gaussian $\mathcal{N}(\boldsymbol{\mu}_I, \boldsymbol{\Sigma}_I)$ as:  
\begin{equation}
    \boldsymbol{\mu}_I = \pi(\boldsymbol{T}_{CW} \boldsymbol{\mu}_W), \quad \boldsymbol{\Sigma}_I = \mathbf{J} \mathbf{W} \boldsymbol{\Sigma}_W \mathbf{W}^\top \mathbf{J}^\top,
\end{equation}
where $\pi$ is the perspective projection, $\mathbf{W}$ is the rotational component of $\boldsymbol{T}_{CW}$, and $\mathbf{J}$ is the Jacobian of the projective approximation.  

Pixel colors $\hat{\mathbf{C}}$ and depths $\hat{D}$ are rendered via front-to-back $\upalpha$-blending:  
\begin{equation}
    \hat{\mathbf{C}} = \sum_{i\in\mathcal{N}} c_i \alpha_i \prod_{j=1}^{i-1}(1-\alpha_j), 
\end{equation}
\begin{equation}
    \hat{D} = \sum_{i\in\mathcal{N}} d_i \alpha_i \prod_{j=1}^{i-1}(1-\alpha_j),
\end{equation}
where $\alpha_i$ is modulated by the projected 2D Gaussian’s density, and $d_i$ is the depth of $G_i$’s centroid in the camera frame. Crucially, this rendering process is fully differentiable, enabling gradient-based optimization of Gaussian parameters.

\paragraph{Map Update}
The 3D Gaussian scene representation is dynamically refined through a continuous cycle of initialization, adaptive density control, and gradient-driven optimization. Initial Gaussians are seeded from sparse geometric cues, such as keyframe-based triangulation or depth estimation, where their initial positions and covariances reflect the uncertainty of sensor observations. To ensure the scene is neither under- nor over-reconstructed, the system employs a density adaptation mechanism: Gaussians are cloned in regions with high photometric reconstruction errors (indicating missing geometry) and pruned if their opacity falls below a minimal threshold or their spatial extent becomes negligible (e.g., due to occlusion or excessive compression). This balance allows the map to grow in complex regions while eliminating redundant elements.

The core of the update process lies in differentiable optimization. We formulate an L1 composite loss function $\mathcal{L}$ to align the rendered outputs with sensor observations, defined as:
\begin{equation}
    \mathcal{L} = \|\hat{\mathbf{C}} - \mathbf{C}_{\text{gt}}\|_1 + \lambda \|\hat{D} - D_{\text{gt}}\|_1,
\end{equation}
where $\hat{\mathbf{C}}$ and $\hat{D}$ denote the rendered RGB image and depth map respectively, while $\mathbf{C}_{\text{gt}}$ and $D_{\text{gt}}$ represent the corresponding ground-truth measurements captured by the sensors. The weighting coefficient $\lambda$ balances the photometric and geometric error terms. Through backpropagation of this differentiable loss function, we jointly optimize a set of 3D Gaussian scene representation parameters including positions $\boldsymbol{\mu}_W$, covariance scales $\mathbf{S}$, rotations $\mathbf{R}$, colors $c$, and opacities $\alpha$. The optimization adjusts Gaussians to better align their projected 2D contributions with observed colors and depths, while implicit regularization maintains physically plausible shapes. Through iterative updates, the Gaussian ensemble converges to a compact yet accurate representation of the scene’s geometry and appearance.

\subsection{Keyframing}
Jointly optimizing all video stream frames with 3D Gaussians and camera poses in real-time is computationally infeasible. Inspired by MonoGS\cite{MonoGS10657715}, we maintain a compact keyframe window \(\mathcal{W}_k\), dynamically curated through inter-frame covisibility analysis. Ideal keyframe management balances two objectives: (1) selecting non-redundant and high-quality frames observing overlapping scene regions to ensure multiview constraints, and (2) maximizing baseline diversity between keyframes to strengthen geometric stability. This strategy minimizes redundancy while preserving reconstruction accuracy, as detailed in our supplementary material.

\paragraph{Selection and Management}
Keyframe selection is governed by a unified scoring function that consolidates three critical criteria: covisibility, baseline span, and motion stability. For each tracked frame $i$, the keyframe score $s_i$ is computed as:  
\begin{equation}
\begin{split}
    s_i &= w_{\text{covis}} \cdot \left(1 - \text{IoU}_{\mathcal{G}}\right) \\&+ w_{\text{base}} \cdot \frac{\| \boldsymbol{t}_{ij} \|}{d_{\text{med}}} \\&- w_{\text{mot}} \cdot \mathbb{I}\left(\boldsymbol{v}_i > v_{\text{max}} \lor \boldsymbol{\omega}_i > \omega_{\text{max}}\right),
\end{split}
\end{equation}
where $\text{IoU}_{\mathcal{G}}$ quantifies the overlap of visible 3D Gaussians between the current frame and the latest keyframe, $\| \boldsymbol{t}_{ij} \| / d_{\text{med}}$ measures the normalized baseline span, and $\boldsymbol{v}_i, \boldsymbol{\omega}_i$ denote linear and angular velocities. The weights $w_{\text{covis}}$, $w_{\text{base}}$, and $w_{\text{mot}}$ balance these factors, while $\mathbb{I}(\cdot)$ enforces motion constraints.  

A frame is selected as a keyframe if $s_i > \tau_{\text{kf}}$ (e.g., $\tau_{\text{kf}} = 0.7$) and the keyframe window $\mathcal{W}_k$ has capacity. To maintain efficiency, redundant keyframes are pruned when their overlap with the latest keyframe exceeds $\tau_{\text{overlap}}$ (e.g., 0.8). This strategy ensures a compact set of non-redundant keyframes that maximize multiview constraints while minimizing computational overhead.

\begin{table*}[ht!]
    \centering
    \resizebox{0.8\linewidth}{!}{
    \setlength{\tabcolsep}{8pt}
    \begin{tabular}{*{6}{c}|*{3}{c}}
        \toprule
        Input & Methods & fr1/desk & fr2/xyz & fr3/office & \textbf{Avg.} & fr1/desk2 & fr1/room & \textbf{Avg.} \\
        \midrule
        \multirow{5.5}{*}{Monocular} 
        & DROID-VO\cite{DROID-SLAM10.5555/3540261.3541527} & 5.12 & 9.88 & 7.30 & 7.43 & - & - & - \\
        \cmidrule(l){2-9}
        & DepthCov-VO\cite{DepthCov10204596} & 5.63 & \textbf{1.20} & 53.4 & 20.08 & - & - & - \\
        \cmidrule(l){2-9}
        & MonoGS\cite{MonoGS10657715} & \underline{3.56} & 4.59 & \underline{3.50} & \underline{3.88} & 77.64 & 79.88 & 79.36 \\
        \cmidrule(l){2-9}
        & Ours & \textbf{1.98} & \underline{3.27} & \textbf{3.14} & \textbf{2.80} & 50.27 & 61.43 & 55.85 \\
        \midrule
        
        \multirow{9}{*}{RGBD}
        & NICE-SLAM\cite{NICE-SLAM9878912} & 4.24 & 6.04 & 3.85 & 4.71 & 4.89 & 33.79 & 19.34 \\
        \cmidrule(l){2-9}
        & Vox-Fusion\cite{Vox-Fusion9995035} & 3.48 & 1.53 & 23.7 & 9.57 & 6.00 & 25.73 & 15.87 \\
        \cmidrule(l){2-9}
        & Point-SLAM\cite{Point-SLAM10377819} & 4.11 & 1.43 & 3.19 & 2.91 & \textbf{4.54} & 33.94 & 19.24 \\
        \cmidrule(l){2-9}
        & SplaTAM\cite{SplaTAM10656349} & 3.34 & \textbf{1.22} & 5.20 & 3.25 & 6.54 & 11.10 & 8.82 \\
        \cmidrule(l){2-9}
        & MonoGS\cite{MonoGS10657715} & \underline{1.59} & 1.36 & \underline{1.55} & \underline{1.50} & 6.30 & \underline{6.64} & \underline{6.47} \\
        \cmidrule(l){2-9}
        & Ours & \textbf{1.34} & \underline{1.26} & \textbf{1.54} & \textbf{1.38} & \underline{4.82} & \textbf{4.66} & \textbf{4.74} \\
        \bottomrule
    \end{tabular}
    }
    \caption{Camera tracking results on TUM for monocular and RGBD(ATE RMSE $\downarrow$ [cm]).}
    \label{table1}
\end{table*}

\paragraph{Gaussian Covisibility}
Covisibility is estimated using the visibility properties of 3D Gaussians during differentiable rendering. A Gaussian $G_k$ is deemed visible in frame $i$ if:  
\begin{itemize}
    \item Its contribution to pixel colors exceeds a minimal threshold ($\alpha$-blending weight $> 0.01$),  
    \item The accumulated opacity along its ray does not surpass 0.5, implicitly handling occlusions. 
\end{itemize}

The covisibility between frames $i$ and $j$ is defined as: 
\begin{equation}
    \text{IoU}_{\mathcal{G}} = \frac{|\mathcal{G}_i \cap \mathcal{G}_j|}{|\mathcal{G}_i \cup \mathcal{G}_j|},
\end{equation} 
where $\mathcal{G}_i$ and $\mathcal{G}_j$ are sets of visible Gaussians. This metric avoids explicit geometric overlap checks and naturally adapts to scene complexity, providing a robust measure of shared scene content.

\paragraph{Motion Data Constraint}
To mitigate motion blur, frames with excessive linear velocity ($\boldsymbol{v}_i > v_{\text{max}}$) or angular velocity ($\boldsymbol{\omega}_i > \omega_{\text{max}}$) are excluded from keyframe candidacy. Thresholds $v_{\text{max}}$ and $\omega_{\text{max}}$ are empirically determined based on sensor characteristics (e.g., camera frame rate and IMU noise). This hard rejection rule ensures only geometrically stable frames are retained, enhancing mapping quality without compromising real-time performance.

\begin{table}[h!]
    \centering
    \resizebox{1.0\linewidth}{!}{
    \setlength{\tabcolsep}{4pt}
    \begin{tabular}{*{6}{c}}
        \toprule
        Methods & Metric & fr1/desk & fr2/xyz & fr3/office & \textbf{Avg.} \\
        \midrule
        \multirow{4}{*}{NICE-SLAM\cite{NICE-SLAM9878912}} 
        & PSNR$\uparrow$ & 13.87 & 17.94 & 15.11 & 15.64 \\
        \cmidrule(l){2-6}
        & SSIM$\uparrow$ & 0.566 & 0.668 & 0.561 & 0.598 \\
        \cmidrule(l){2-6}
        & LPIPS$\downarrow$ & 0.485 & 0.327 & 0.382 & 0.398 \\
        \midrule
        \multirow{4}{*}{Vox-Fusion\cite{Vox-Fusion9995035}} 
        & PSNR$\uparrow$ & 15.79 & 16.53 & 17.22 & 16.51 \\
        \cmidrule(l){2-6}
        & SSIM$\uparrow$ & 0.653 & 0.711 & 0.677 & 0.68 \\
        \cmidrule(l){2-6}
        & LPIPS$\downarrow$ & 0.514 & 0.423 & 0.459 & 0.465 \\
        \midrule
        \multirow{4}{*}{Point-SLAM\cite{Point-SLAM10377819}} 
        & PSNR$\uparrow$ & 13.87 & 17.61 & 18.93 & 16.8 \\
        \cmidrule(l){2-6}
        & SSIM$\uparrow$ & 0.627 & 0.715 & 0.744 & 0.695 \\
        \cmidrule(l){2-6}
        & LPIPS$\downarrow$ & 0.564 & 0.562 & 0.442 & 0.523 \\
        \midrule
        \multirow{4}{*}{SplaTAM\cite{SplaTAM10656349}} 
        & PSNR$\uparrow$ & \underline{22.63} & 24.55 & 22.71 & 23.29 \\
        \cmidrule(l){2-6}
        & SSIM$\uparrow$ & \textbf{0.852} & \textbf{0.935} & \underline{0.876} & \textbf{0.888} \\
        \cmidrule(l){2-6}
        & LPIPS$\downarrow$ & \underline{0.239} & \textbf{0.103} & 0.221 & \textbf{0.188} \\
        \midrule
        \multirow{4}{*}{MonoGS\cite{MonoGS10657715}} 
        & PSNR$\uparrow$ & 22.56 & \underline{24.86} & \underline{24.37} & \underline{23.93} \\
        \cmidrule(l){2-6}
        & SSIM$\uparrow$ & 0.774 & 0.8 & 0.823 & 0.799 \\
        \cmidrule(l){2-6}
        & LPIPS$\downarrow$ & 0.247 & 0.211 & \underline{0.21} & 0.223 \\
        \midrule
        \multirow{4}{*}{Ours} 
        & PSNR$\uparrow$ & \textbf{23.98} & \textbf{25.37} & \underline{24.29} & \textbf{24.55} \\
        \cmidrule(l){2-6}
        & SSIM$\uparrow$ & \underline{0.833} & \underline{0.851} & \textbf{0.881} & \underline{0.855} \\
        \cmidrule(l){2-6}
        &LPIPS$\downarrow$ & \textbf{0.209} & \underline{0.191} & \textbf{0.196} & \underline{0.199} \\
        \bottomrule
    \end{tabular}
    }
    \caption{Rendering performance on TUM for RGBD.}
    \label{table2}
\end{table}

\section{Experiments}
\label{sec:experiments}

We conducted extensive experiments to evaluate the performance of GI-SLAM in terms of both camera tracking accuracy and mapping quality. Our experiments were designed to benchmark against state-of-the-art methods and assess the effectiveness of GI-SLAM in diverse settings, including different datasets and evaluation metrics.

\subsection{Experimental Setup}

\paragraph{Implementation details}
GI-SLAM is implemented in Python using the PyTorch framework, incorporating CUDA code for Gaussian splatting. The training and evaluation were performed on a desktop PC equipped with a 6.0GHz Intel Core i9-14900K CPU and an NVIDIA RTX 4090 GPU. More technical details can be found in the supplemental materials.

\paragraph{Datasets}
To evaluate the performance of GI-SLAM, we conducted experiments on the EuRoC\cite{euroc10.1177/0278364915620033} and TUM-RGBD\cite{tum6385773} datasets. The EuRoC dataset contains stereo vision data and IMU readings, making it suitable for evaluating the model in a stereo + IMU configuration. Although the TUM-RGBD dataset only includes accelerometer data, it is a standard benchmark in SLAM research. Therefore, we ensured compatibility with IMU data containing only accelerometer information to validate our model under monocular + IMU and RGBD + IMU settings.

\paragraph{Metric}
For camera tracking accuracy, we employed the root mean square error (RMSE) of the Absolute Trajectory Error (ATE) calculated on keyframes. To assess mapping quality, we followed the photometric rendering quality metrics used in MonoGS\cite{MonoGS10657715}, including PSNR, SSIM, and LPIPS. These metrics were computed every five frames, excluding keyframes (training viewpoints). 

\paragraph{Baselines}
We compared GI-SLAM against state-of-the-art open-source methods in the 3DGS SLAM domain, including MonoGS\cite{MonoGS10657715} and SplaTAM\cite{SplaTAM10656349}. Additionally, we benchmarked against earlier learning-based methods and NeRF-based approaches. For camera tracking accuracy, comparisons were made with advanced learning-based direct visual odometry (VO) methods and 3DGS SLAM methods, such as DepthCov\cite{DepthCov10204596}, DROID-SLAM\cite{DROID-SLAM10.5555/3540261.3541527}, SplaTAM\cite{SplaTAM10656349} and MonoGS\cite{MonoGS10657715}. For mapping quality, comparisons were conducted with current leading NeRF-based SLAM methods\cite{NICE-SLAM9878912,Vox-Fusion9995035,Point-SLAM10377819} and 3DGS SLAM approache\cite{SplaTAM10656349,MonoGS10657715}.

\subsection{Results \& Discussion}

\paragraph{Camera Tracking Results}
We evaluated the camera tracking performance of GI-SLAM on both the TUM and EuRoC datasets. On TUM, experiments were conducted under monocular and RGBD settings, while on EuRoC a stereo configuration was employed. As shown in \mbox{\cref{table1}} (TUM results for monocular and RGBD) and \mbox{\cref{table5}}  (EuRoC stereo results), GI-SLAM consistently achieves lower RMSE values for ATE on keyframes compared to current state-of-the-art approaches. In particular, the integration of our IMU loss function plays a significant role in stabilizing pose estimates under challenging conditions. For instance, in the monocular setup our approach reduces the trajectory error over the prior SOTA baseline\cite{MonoGS10657715} by more than 20\% from 3.88cm to 2.80cm.

The benefits of fusing IMU data become more apparent in scenarios with rapid camera motion or low-texture environments, where visual information alone might be insufficient. As illustrated in \mbox{\cref{fig2}}, our approach enables the camera tracking to achieve higher accuracy at turns. The inclusion of inertial data not only improves accuracy but also enhances the robustness of the tracking process over longer sequences, effectively mitigating drift. These improvements underscore the importance of our proposed IMU integration strategy, as reflected in the quantitative results.

\paragraph{Rendering Quality Results}
Mapping quality was assessed using photometric rendering metrics such as PSNR, SSIM, and LPIPS, computed on non-keyframe views at regular intervals. \mbox{\cref{table2}} summarize the performance on the TUM dataset for RGBD setup. GI-SLAM demonstrates superior rendering performance, with notable improvements in PSNR and SSIM values and a corresponding decrease in LPIPS scores compared to the baseline methods. For example, when compared to MonoGS\cite{MonoGS10657715}, the PSNR improved from 23.93 to 24.55, with similar trends observed in SSIM and LPIPS metrics.

These gains can be largely attributed to our novel keyframe selection strategy based on motion constraint and Gaussian covisibility. By ensuring that only frames with minimal motion blur are selected for mapping, our method produces reconstructions that are both visually consistent and photorealistic. The improvements in rendering quality metrics indicate that our approach effectively reduces artifacts and preserves fine scene details, thus providing a more reliable basis for downstream tasks that depend on accurate scene representations.

\begin{table}[h!]
    \centering
    \resizebox{0.8\linewidth}{!}{
    \setlength{\tabcolsep}{8pt}
    \begin{tabular}{*{5}{c}}
        \toprule
        Input & IMU & fr1/desk & fr2/xyz & Avg. \\
        \midrule
        \multirow{2.5}{*}{Monocular}
        & w/ & 1.98 & 3.27 & 2.63 \\
        \cmidrule(l){2-5}
        & w/o & 3.51 & 4.29 & 3.90 \\
        \midrule
        \multirow{2.5}{*}{RGBD}
        & w/ & 1.34 & 1.26 & 1.30 \\
        \cmidrule(l){2-5}
        & w/o & 1.55 & 1.34 & 1.45 \\
        \bottomrule
    \end{tabular}
    }
    \caption{Ablation Study with/without IMU on the TUM(ATE RMSE $\downarrow$ [cm]).}
    \label{table3}
\end{table}

\paragraph{Ablative Analysis}
To further validate the contributions of our two main innovations, we performed ablative studies by selectively disabling each component. First, when the IMU fusion module was removed, the camera tracking performance deteriorated noticeably. The ATE RMSE increased from 2.63 cm (with IMU) to 3.90 cm (without IMU) under monocular conditions, and from 1.30 cm to 1.45 cm in RGBD configurations, as detailed in  \mbox{\cref{table3}}. This result confirms that the IMU loss function is essential for enhancing the stability and accuracy of pose estimation, particularly in dynamic or visually challenging scenarios.

\begin{table}[h!]
    \centering
    \resizebox{1.0\linewidth}{!}{
    \setlength{\tabcolsep}{8pt}
    \begin{tabular}{*{6}{c}}
        \toprule
        \makecell{Motion \\ constraint} & Metric & fr1/desk & fr2/xyz & fr3/office & \textbf{Avg.} \\
        \midrule
        \multirow{4}{*}{w/} 
        & PSNR$\uparrow$ & 23.98 & 25.37 & 24.29 & 24.55 \\
        \cmidrule(l){2-6}
        & SSIM$\uparrow$ & 0.833 & 0.851 & 0.881 & 0.855 \\
        \cmidrule(l){2-6}
        &LPIPS$\downarrow$ & 0.209 & 0.191 & 0.196 & 0.199 \\
        \midrule
        \multirow{4}{*}{w/o} 
        & PSNR$\uparrow$ & 22.91 & 24.27 & 24.15 & 23.78 \\
        \cmidrule(l){2-6}
        & SSIM$\uparrow$ & 0.802 & 0.813 & 0.839 & 0.818 \\
        \cmidrule(l){2-6}
        &LPIPS$\downarrow$ & 0.225 & 0.197 & 0.212 & 0.211 \\
        \bottomrule
    \end{tabular}
    }
    \caption{Ablation study of rendering on TUM with/without motion constraint as a factor in keyframe selection.}
    \label{table4}
\end{table}

\begin{table}[h!]
    \centering
    \resizebox{0.8\linewidth}{!}{
    \setlength{\tabcolsep}{8pt}
    \begin{tabular}{*{6}{c}}
        \toprule
        Methods & mh01 & mh02 & v101 & Avg. \\
        \midrule
        MonoGS\cite{MonoGS10657715} & 13.09 & 8.24 & 9.73 & 19.35 \\
        \midrule
        Ours & 9.77 & 6.91 & 6.81 & 7.83 \\
        \bottomrule
    \end{tabular}
    }
    \caption{Camera tracking results on 3 easy sequences of EuRoC(ATE RMSE $\downarrow$ [cm]).}
    \label{table5}
\end{table}

In a separate experiment, we replaced our motion-constrained keyframe selection strategy with a conventional selection mechanism that does not account for motion blur. This modification led to a decline in mapping quality: PSNR decreased from 24.55 to 23.78, SSIM dropped from 0.855 to 0.818, and LPIPS increased from 0.199 to 0.211, as shown in \mbox{\cref{table4}}. These findings emphasize that careful keyframe selection is crucial for achieving high-quality map reconstructions, as it effectively filters out frames that could introduce noise and artifacts due to motion blur.

The ablative analysis highlights that both the IMU integration and the advanced keyframe selection strategy are vital components of GI-SLAM. Their combined effect not only improves camera tracking accuracy but also significantly enhances the rendering quality of the generated maps.

In summary, the experimental results demonstrate the effectiveness and robustness of GI-SLAM across various sensor configurations and datasets. The integration of IMU data via a dedicated loss function, coupled with an intelligent keyframe selection strategy, contributes to significant improvements in both tracking accuracy and mapping quality. These findings validate our design choices and suggest that GI-SLAM is well-suited for real-world 3DGS SLAM applications.

\section{Conclusion}
\label{sec:conclusion}

In this paper, we proposed GI-SLAM, a novel 3DGS SLAM framework that integrates IMU data through a dedicated loss function and employs a motion-constrained keyframe selection strategy based on Gaussian co-visibility, leading to improved camera tracking accuracy and mapping quality as demonstrated on the TUM and EuRoC datasets. However, our approach does not explicitly address the noise inherent in IMU measurements, and the metric scale ambiguity in monocular SLAM configurations remains unresolved. We plan to tackle these challenges in future work to further enhance the robustness and applicability of GI-SLAM.

{
    \small
    \bibliographystyle{ieeenat_fullname}
    \bibliography{main}
}

\end{document}